\documentclass[sigconf]{acmart}

\usepackage{balance}
\usepackage{titlesec}
\usepackage{mathrsfs}

\usepackage{amsmath,amssymb,amsfonts,amsthm}
\usepackage{enumitem} 
\usepackage{graphicx}
\usepackage{extarrows}
\usepackage{subfigure}
\usepackage{verbatim}
\usepackage{makecell}
\usepackage{diagbox}
\usepackage{bm}
\usepackage{multirow}
\usepackage{float}
\usepackage{placeins}
\usepackage[linesnumbered, ruled]{algorithm2e}
\SetAlFnt{\small}
\usepackage{epstopdf}
\usepackage{setspace}
\usepackage[super]{nth}
\usepackage{fancyhdr}
\usepackage{slashbox}
\usepackage{lipsum,multicol}
\usepackage{url}
\usepackage{xcolor}
\usepackage{booktabs}
\usepackage{caption}
\usepackage{ragged2e}
\usepackage{wrapfig}
\usepackage{tikz}
\usepackage{hyperref}
\usepackage{colortbl}
\usepackage{tcolorbox} 
\usepackage[normalem]{ulem}

\def\BibTeX{{\rm B\kern-.05em{\sc i\kern-.025em b}\kern-.08em
    T\kern-.1667em\lower.7ex\hbox{E}\kern-.125emX}}

\SetCommentSty{mycommfont}

\newcommand{\model}{RouteGoT\xspace}
\allowdisplaybreaks[4]

\begin{document}

\title{\model: Node-Adaptive Routing for Cost-Efficient Graph of Thoughts Reasoning} 

\author{Yuhang Liu}
\authornote{Both authors contributed equally to this research.} 
\affiliation{%
  \institution{School of New Media and Communication, Tianjin University}
  \city{Tianjin}
  \country{China}
}
\email{liuyuhang_13@tju.edu.cn}

\author{Ruijie Wang}
\authornotemark[1] 
\affiliation{%
  \institution{School of Computer Science and Engineering, Beihang University}
  \city{Beijing}
  \country{China}}
\email{ruijiew@buaa.edu.cn}

\author{Yunlong Chu}
\authornotemark[1] 
\affiliation{%
  \institution{School of New Media and Communication, Tianjin University}
  \city{Tianjin}
  \country{China}}
\email{2024245030@tju.edu.cn}

\author{Bing Hao}
\affiliation{%
  \institution{School of New Media and Communication, Tianjin University}
  \city{Tianjin}
  \country{China}}
\email{haobing@tju.edu.cn}

\author{Yumeng Lin}
\affiliation{%
  \institution{School of New Media and Communication, Tianjin University}
  \city{Tianjin}
  \country{China}}
\email{lym619@tju.edu.cn}

\author{Shengzhong Liu}
\affiliation{%
  \institution{Shanghai Jiao Tong University}
  \city{Shanghai}
  \country{China}}
\email{shengzhong@sjtu.edu.cn}

\author{Minglai Shao}
\authornote{Corresponding author.} 
\affiliation{%
  \institution{School of New Media and Communication, Tianjin University}
  \city{Tianjin}
  \country{China}}
\email{shaoml@tju.edu.cn}


\begin{abstract}
Large Language Models (LLMs) excel at multi-step reasoning, yet increasing the structural complexity of inference does not consistently improve \emph{system-level} returns. Methods such as Tree of Thoughts (ToT), Graph of Thoughts (GoT), and Adaptive Graph of Thoughts (AGoT) can boost accuracy on some benchmarks, but often introduce substantial overhead in token consumption and latency, and their gains can be unstable across task distributions---sometimes underperforming simpler Chain-of-Thought (CoT) or direct input-output prompting (IO).
We attribute this inefficiency to \emph{stage-wise and node-wise heterogeneity} inside GoT-style reasoning pipelines: high-quality planning and final synthesis are globally coupled and typically benefit from strong models, whereas many intermediate subtasks are localized and can be solved accurately by lighter models with far fewer tokens. Motivated by these observations, we propose \model, a budget-controllable, node-adaptive routing framework for graph-structured reasoning. \model performs \emph{in-graph} routing by prioritizing strong models for planning and synthesis, while dynamically allocating lightweight models and cost-effective strategies to leaf subtasks based on predicted difficulty. It further integrates explicit budget constraints into a global inference scheduler to control graph expansion (depth/branching) under a user-specified token budget, enabling predictable performance--cost trade-offs. Experiments across reasoning, retrieval, and multi-hop QA benchmarks show that \model matching or improving accuracy while substantially reducing token usage; specifically, it achieves an average 8.1 percentage points accuracy improvement and 79.1\% output token reduction compared to AGoT. Furthermore, RouteGoT outperforms existing routing baselines by maintaining a superior cost--accuracy trade-off, demonstrating improved robustness under varying budget targets and task distributions.
\end{abstract}


%

\maketitle

\renewcommand{\shortauthors}{xxx et al.}

\section{Introduction}
\label{sec:intro}

Large Language Models (LLMs) have demonstrated strong capabilities in mathematical reasoning, open-domain question answering, and multi-step decision making in data-intensive analytics and knowledge discovery pipelines~\cite{lu-etal-2023-survey,chen2025reasoningerasurveylong,yue2025surveylargelanguagemodel,zhang-etal-2023-survey-efficient,wang2025surveylargelanguagemodels,xia-etal-2025-beyond}. This success has motivated a shift from linear prompting to structured reasoning frameworks. Chain-of-Thought (CoT)~\cite{cot} improves problem solving by explicitly generating intermediate reasoning steps. More recently, structured inference methods with explicit topologies, including Tree of Thoughts (ToT)~\cite{tot}, Graph of Thoughts (GoT)~\cite{got}, and Adaptive Graph of Thoughts (AGoT)~\cite{agot}, incorporate decomposition, search, and multi-stage aggregation, enabling non-linear reasoning paths and self-correction~\cite{reward,chen2024plan,Besta_2025}.

\begin{figure}[t]
  \centering
  \includegraphics[width=\linewidth]{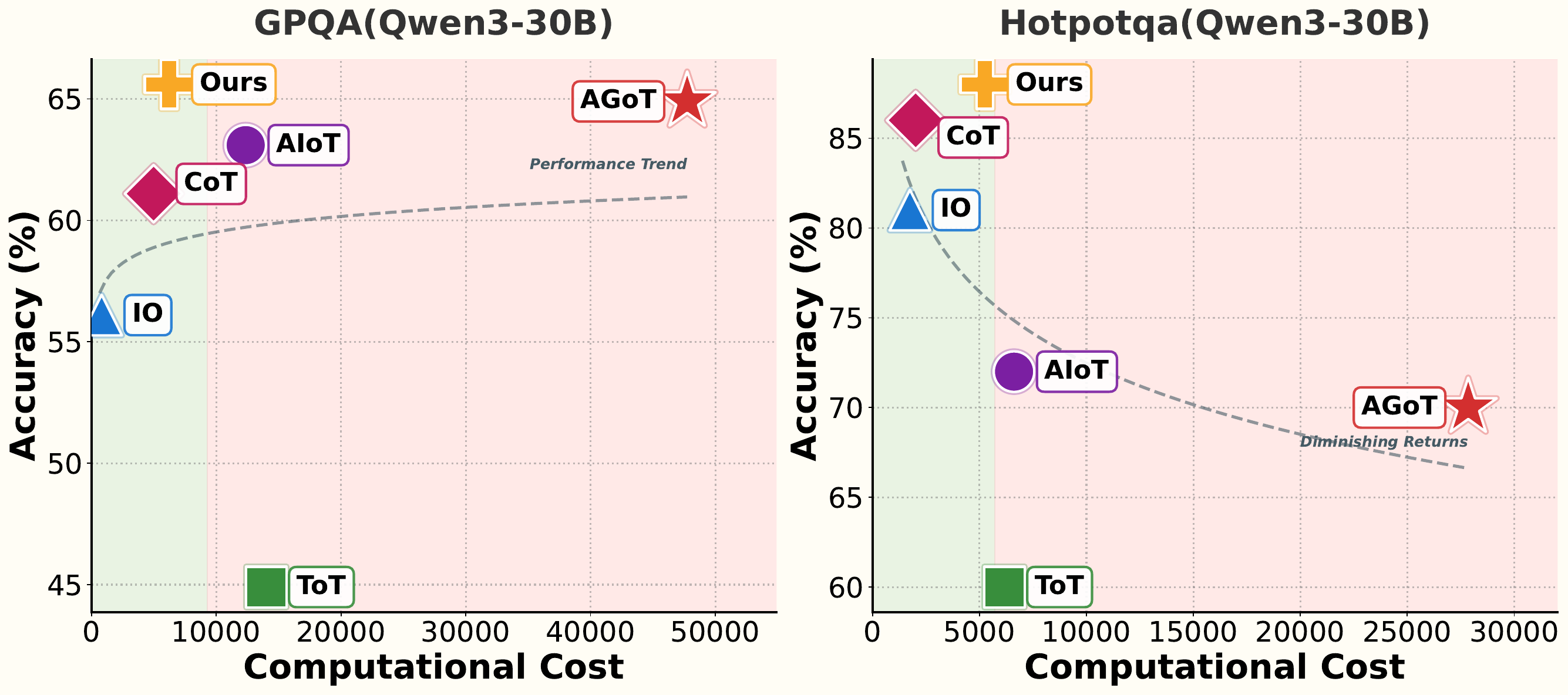}
  \caption{Cost-Accuracy Trade-off. Background colors distinguish low-cost (light green) from high-cost (light red) regions based on the median token consumption across all methods.}
  \label{fig:cost_accuracy}
\end{figure}

In this work, we focus on graph-structured reasoning systems, such as GoT and AGoT, which typically follow a pipeline of planning $\rightarrow$ subtask execution $\rightarrow$ global synthesis. While this paradigm improves compositionality, it creates significant uncertainty regarding end-to-end inference costs in production. Factors such as decomposition depth, branching width, and subtask length can vary dramatically across inputs, causing high variance in both token consumption and latency. This makes computational budgets difficult to predict or enforce, often leading to resource over-provisioning or unexpected latency spikes. More importantly, empirical evidence suggests that higher compute does not reliably yield higher accuracy. As shown in Figure~\ref{fig:cost_accuracy}, while AGoT achieves strong accuracy on GPQA, it does so at a prohibitive token cost; conversely, on HotpotQA, it consumes substantially more tokens than CoT or Direct IO yet performs worse. Such inconsistency challenges the assumption that additional computation necessarily yields higher accuracy~\cite{tokenprune,be3r,lookahead}.

A key reason for this inefficiency is that node difficulty varies widely within a single reasoning graph. Global steps like planning and final synthesis are coupled and often require capable models to ensure decomposition quality and aggregation consistency~\cite{tale}. In contrast, many intermediate subtasks are local and significantly simpler than the original problem. While smaller models can often solve these subtasks using far fewer tokens, standard GoT and AGoT pipelines typically process all nodes with a uniform heavyweight model, creating substantial redundancy. This observation aligns with recent work on cost-aware routing~\cite{chen2025harnessing}, but existing mechanisms are usually task-level: they select a model or strategy at the entry point and keep it fixed at inference~\cite{zhuang2025embedllm,rtr,ong2025routellm,chen2025concur}. This approach treats the reasoning graph as a monolithic entity, ignoring internal heterogeneity. Selecting a strong model globally wastes computation on trivial nodes, while choosing a weak model risks failure on critical nodes, compromising the entire reasoning chain.

Based on these observations, we identify two limitations in current efficient GoT systems:
\textbf{(1) Budget Controllability:} under a user-specified budget, the system should explicitly constrain graph expansion (depth/width) and allocate computation predictably, without extensive tuning that fails to generalize across distributions.
\textbf{(2) In-process Orchestration:} the system should make \emph{dynamic} decisions during reasoning, exploiting node-level heterogeneity to reduce redundancy on simple subtasks while preserving quality at critical stages.

To address these issues, we propose \model, a node-aware routing framework embedded \emph{inside} GoT-style inference. RouteGoT routes each leaf node to an appropriate model-bound action based on predicted success and difficulty, and employs a global budget scheduler to enforce strict budget adherence while controlling graph expansion. Across reasoning, retrieval, and multi-hop QA benchmarks, RouteGoT achieves an average 8.1 percentage-point accuracy improvement and a 79.1\% output token reduction compared to AGoT. It also outperforms existing routing baselines by maintaining a stronger cost--accuracy trade-off, with improved robustness under varying budget targets and task distributions, making it suitable for real-world production environments.

Our contributions are:
\begin{itemize}[leftmargin=8pt]
    \item We empirically analyze cost variance and cost--accuracy instability in AGoT-style inference, showing that higher token cost does not guarantee higher accuracy.
    \item We propose RouteGoT, a node-adaptive routing framework that performs \emph{in-graph} model/strategy allocation and integrates explicit budget control for predictable graph expansion.
    \item We demonstrate substantial token savings and improved robustness on multiple benchmarks under different budget targets.
\end{itemize}

\section{Problem Formulation}
\label{sec:2_background}

Given an input question $x$, we perform GoT-style reasoning by constructing a directed reasoning graph $G=(V,E)$, where each node $v\in V$ corresponds to a (sub)task and edges encode dependencies. Inference follows three stages: \emph{planning} $\rightarrow$ \emph{subtask execution} $\rightarrow$ \emph{global synthesis}. RouteGoT assumes every node can be represented as a standalone ``task instance'' through a unified input template (Section~\ref{sec:node_input}), so the same router can be applied to both the root task and intermediate subtasks.

\noindent \textbf{Token cost and global budget.}
We measure inference cost in tokens. Let $\text{call}_m$ denote the $m$-th model invocation and $M(x)$ the total number of calls. The total cost is
\begin{equation}
C(x)=\sum_{m=1}^{M(x)} \mathrm{tokens}\big(\text{call}_m\big).
\end{equation}
The system is configured with a user-specified global token budget $B_{\text{total}}$. We also reserve a synthesis buffer $B_{\text{syn}}$ (Section~\ref{sec:inference_scheduler}) to guarantee the final aggregation step.

\noindent \textbf{Node-level constrained decision.}
For each pending leaf node $v$, the router selects an action $a(v)$ from a fixed action set $\mathcal{A}$. Each action has (instance-dependent) success probability $\mathbb{P}(\text{success}\mid v,a)$ and token cost. Since costs---especially for graph expansion---cannot be known exactly before execution, RouteGoT uses an \emph{estimated} per-action cost $\hat{c}_a(v)$.
Let $B_{\text{rem}}$ denote the remaining budget after accounting for accumulated cost and synthesis reserve, and let $B_{\text{cap}}(v)$ be a per-node budget cap derived from a predicted difficulty budget (Section~\ref{sec:tier_predictor}). We define an \emph{effective} node budget bound:
\begin{equation}
B(v)=\min\big(B_{\text{rem}},\,B_{\text{cap}}(v)\big).
\end{equation}
The router targets the following constrained decision at each node:
\begin{equation}
a^\star(v)=\arg\max_{a\in\mathcal{A}}\ \mathbb{P}(\text{success}\mid v,a)
\quad \text{s.t.}\quad \hat{c}_a(v)\le B(v).
\label{eq:constrained_obj}
\end{equation}
This captures our goal: allocate expensive computation only when needed, while ensuring decisions remain feasible under both the global budget and the node-level compute regime.

\begin{figure*}[t]
  \centering
  \includegraphics[width=0.9\textwidth]{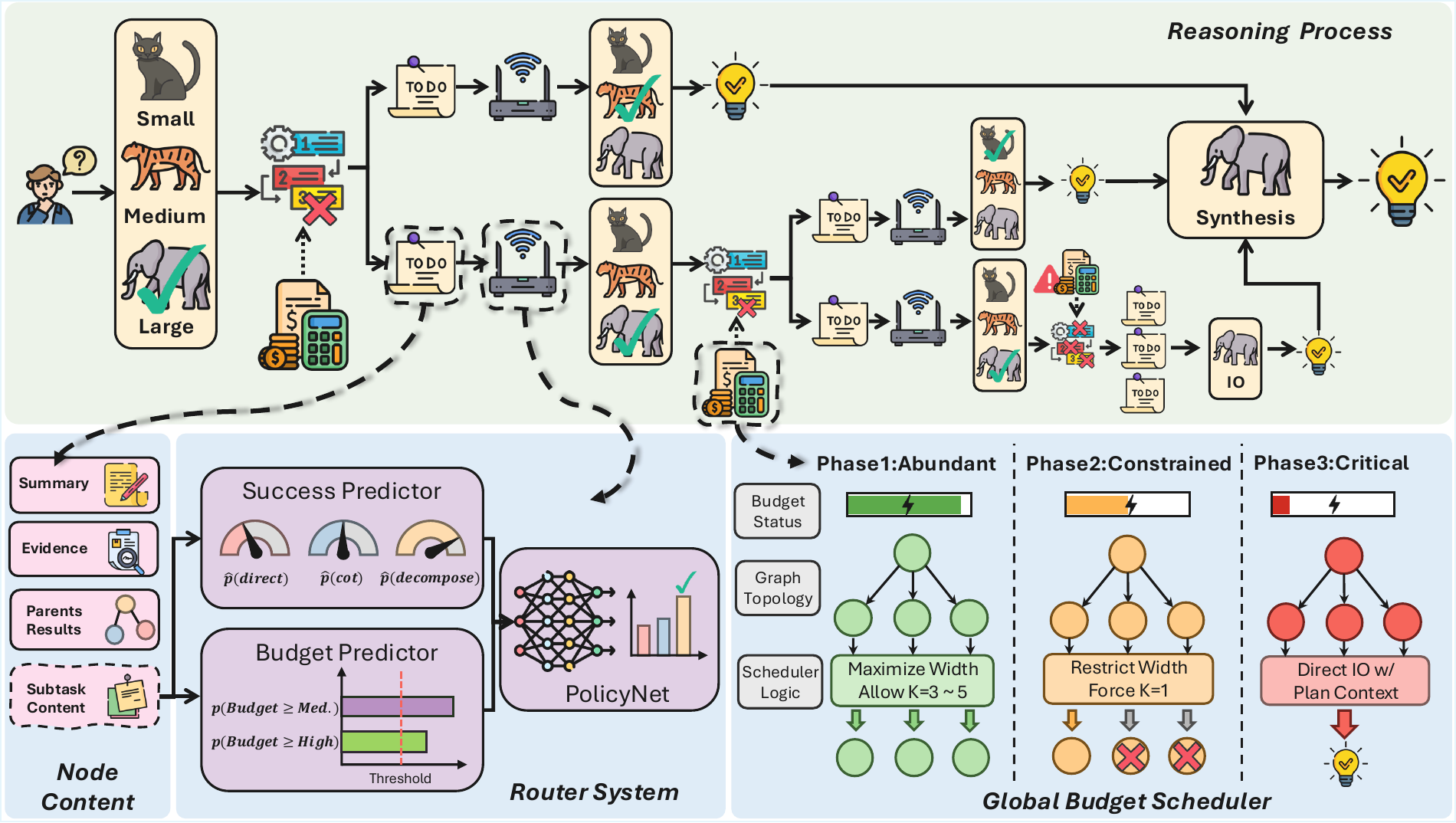}
  \captionsetup{skip=2pt}
  \caption{Overview of the \model framework.}
  \label{fig:framework}
\end{figure*}

\section{The \model Framework}
\label{sec:3_framework}

\subsection{Overview}
RouteGoT is a node-adaptive routing framework integrated into GoT-style inference. It is motivated by the observation that node difficulty is highly uneven within a single reasoning graph: a small set of globally coupled steps, such as planning and synthesis, typically requires a strong model, whereas many intermediate subtasks are local and can be handled by smaller models with far lower cost. RouteGoT therefore makes node-level action choices under budget constraints using three learned modules: a Success Predictor that estimates per-action success likelihood, a Budget Predictor that assigns a coarse per-node difficulty budget, and a PolicyNet that maps these signals to a budget-conditioned action distribution. A Global Budget Scheduler then enforces the overall token limit $B_{\text{total}}$ and regulates graph expansion (Figure~\ref{fig:framework}).

\subsection{\model\ Reasoning Pipeline}
RouteGoT runs GoT-style inference by building a dynamic graph in which each node corresponds to a concrete reasoning step or sub-problem. As the graph grows, the router makes per-node decisions and the executor carries them out, so compute is allocated based on the node content and the current budget state. Appendix~\ref{sec:appendix_algorithm} provides full pseudocode for the end-to-end procedure.

\subsubsection{Action Space with Model Binding}
For each pending leaf node $v$, the router selects one action:
\begin{equation}
a(v)\in\{\textsf{IO},\ \textsf{CoT},\ \textsf{Decompose}\}.
\end{equation}
Each action is tied to a model scale and prompting style: \textsf{IO} uses a small model for direct answering, \textsf{CoT} uses a medium model with chain-of-thought prompting, and \textsf{Decompose} uses a large model to expand a subgraph and produce a branch summary.

\noindent \textbf{Macro-action semantics of \textsf{Decompose}.}
\textsf{Decompose} acts as a macro-action that expands the current node by proposing child subtasks and a branch summary, after which the children are routed and executed recursively. Since the resulting cost depends on downstream routing choices and the induced topology, RouteGoT estimates expansion cost online and applies a plan-guided fallback when needed (Section~\ref{sec:inference_scheduler}).

\subsubsection{Node Input Packing: Summary, Evidence, Parent Results, Subtask}
\label{sec:node_input}
Each node $v$ is formatted as a standalone task instance with four fields: a compact summary $s(v)$, planner-provided evidence $e(v)$, parent results $r(\mathrm{pa}(v))$, and the subtask content $c(v)$:
\begin{equation}
\mathrm{inp}(v)=\mathrm{Template}\big(s(v),\, e(v),\, r(\mathrm{pa}(v)),\, c(v)\big).
\end{equation}
The summary carries global or branch context without repeatedly inserting the full question, evidence captures local constraints, and parent results pass intermediate outputs.

\subsubsection{Root Initialization, Branch Summaries, and Global Synthesis}
The root calls \textsf{Decompose} once to initialize the graph, producing $K\in[1,5]$ subtasks and a root summary $s_{\text{root}}$ that all first-layer nodes inherit. When a non-root node takes \textsf{Decompose}, the large model generates child subtasks and a branch summary $s_{\text{branch}}$, which downstream nodes inherit to keep context compact and branch-specific. Execution ends when all leaves are solved or expansion stops under budget or depth limits, after which a large model performs a single global synthesis over the available node results to produce $\hat{y}(x)$. If some queued nodes remain unsolved due to budget exhaustion, we include an explicit \texttt{UNSOLVED\_DUE\_TO\_BUDGET} marker in the synthesis input so the model does not implicitly invent missing results.

\subsection{Training Data and Supervision}
\label{sec:data_supervision}
RouteGoT does \emph{not} require node-level ground-truth labels for intermediate subtasks at inference time. Instead, we train the router offline using a \emph{three-strategy contrast pool} constructed from standard task datasets with reference answers.

\noindent For each training instance $u$ (a standalone task with a ground-truth answer), we run all three strategies $a\in\{\textsf{IO},\textsf{CoT},\textsf{Decompose}\}$ and log (i) correctness $y_a(u)\in\{0,1\}$ by matching the strategy output against the ground truth and (ii) the realized token cost $c_a(u)\in\mathbb{R}^+$ from execution traces. At inference time, we format every graph node as a task instance $\mathrm{inp}(v)$ and apply the same router to both the root and intermediate subtask nodes.

\noindent \textbf{Cost definition.}
For \textsf{IO} and \textsf{CoT}, $c_a(u)$ is the total tokens of a single model call (input+output). For \textsf{Decompose}, which triggers a graph reasoning procedure, we define cost as the total tokens accumulated over the entire graph run:
\begin{equation}
c_{\textsf{Decompose}}(u)=\sum_{\text{call}\in \text{GraphRuns}(u)} \mathrm{tokens}(\text{call}).
\end{equation}

\noindent \textbf{All-fail handling.}
Instances with $y_{\textsf{IO}}(u)=y_{\textsf{CoT}}(u)=y_{\textsf{Decompose}}(u)=0$ are treated as \emph{all-fail} and excluded when fitting budget thresholds to reduce skew from outliers.

\subsection{In-Graph Router}
RouteGoT approximates the constrained choice in Eq.~\eqref{eq:constrained_obj} with three learned modules: a Success Predictor, a Budget Predictor, and a budget-conditioned PolicyNet.

\subsubsection{Success Predictor: Multi-head Binary Success Estimation}
\label{sec:success_predictor}
The Success Predictor estimates action-wise success probabilities:
\begin{equation}
\hat{p}_a(u)=\mathbb{P}(y_a(u)=1\mid u),\quad a\in\{\textsf{IO},\textsf{CoT},\textsf{Decompose}\}.
\end{equation}
We use a multi-head binary setup because more than one action can be correct for the same instance. With logits $f_a(u)$ and $\hat{p}_a(u)=\sigma(f_a(u))$, we minimize the multi-head binary cross-entropy:
\begin{equation}
\mathcal{L}_{\text{sp}}(u)=\sum_{a}\mathrm{BCE}\big(y_a(u),\hat{p}_a(u)\big).
\end{equation}
We optionally add a margin ranking term so that correct actions score above incorrect ones:
\begin{equation}
\mathcal{L}_{\text{rank}}(u)=
\sum_{a^+\in\mathcal{A}^+(u)}\sum_{a^-\in\mathcal{A}^-(u)}
\max\big(0,\ m-(f_{a^+}(u)-f_{a^-}(u))\big),
\end{equation}
where $\mathcal{A}^+(u)=\{a:\ y_a(u)=1\}$ and $\mathcal{A}^-(u)=\{a:\ y_a(u)=0\}$.
The overall objective is
\begin{equation}
\min_{\theta}\ \mathbb{E}_u\big[\mathcal{L}_{\text{sp}}(u)+\lambda_{\text{rank}}\mathcal{L}_{\text{rank}}(u)\big].
\end{equation}
At inference, we pass $\mathrm{logit}(\hat{p}_a)$ to reduce saturation effects.

\subsubsection{Budget Predictor: Ordinal Difficulty Budget Prediction}
\label{sec:tier_predictor}
Exact token costs are noisy, especially when an action expands the graph. Instead of regressing costs directly, we predict a stable difficulty budget that approximates the minimum compute regime needed.

\noindent \textbf{Budget construction.}
For training, we define the required cost as $c_{\text{req}}(u)=\min_{a:\ y_a(u)=1} c_a(u)$, excluding all-fail cases.
Using global thresholds $B_{25}$ and $B_{75}$ as the 25th and 75th percentiles of $c_{\text{req}}$, we assign a 3-budget label $t(u)\in\{0,1,2\}$:
\begin{equation}
t(u)= \begin{cases}
0, & c_{\text{req}}(u)\le B_{25},\\
1, & B_{25} < c_{\text{req}}(u)\le B_{75},\\
2, & c_{\text{req}}(u) > B_{75}.
\end{cases}
\end{equation}
These labels map to per-node caps $(B_0,B_1,B_2)=(B_{25},B_{75},+\infty)$, where the top is left unbounded so the feasible set is never empty.

\noindent \textbf{Ordinal training.}
We train with ordinal regression on $z_k(u)=\mathbb{I}[t(u)\ge k]$ for $k\in\{1,2\}$.
The model predicts logits $g_k(u)$ with $\hat{q}_k(u)=\sigma(g_k(u))$ and minimizes
\begin{equation}
\mathcal{L}_{\text{cap}}(u)=w(t(u))\sum_{k=1}^2 \mathrm{BCE}\big(z_k(u),\hat{q}_k(u)\big),
\end{equation}
where $w(2)>w(1)\ge w(0)$ penalizes under-estimation that would exclude needed compute.
At inference, we decode budgets via thresholds $\tau_k$: $\hat{t}(u)=\sum_k \mathbb{I}[\hat{q}_k(u)\ge\tau_k]$.

\subsubsection{PolicyNet: Budget-Conditioned Constrained Action Policy}
\label{sec:policynet}
PolicyNet outputs a budget-conditioned action distribution, approximating Eq.~\eqref{eq:constrained_obj} under noisy predictors. Its input is
\begin{equation}
\mathbf{z}(u)=\Big[ \begin{aligned}
&\mathrm{logit}(\hat{p}_{\textsf{IO}}(u)),\ \mathrm{logit}(\hat{p}_{\textsf{CoT}}(u)),\\
&\mathrm{logit}(\hat{p}_{\textsf{Decompose}}(u)),\ \mathrm{onehot}(\hat{t}(u))
\end{aligned} \Big],
\end{equation}
and the resulting policy is
\begin{equation}
\pi_\phi(a\mid u)=\mathrm{softmax}\big(h_\phi(\mathbf{z}(u))\big).
\end{equation}

\noindent \textbf{Feasible-set supervision.}
To train PolicyNet under budgets, we construct a feasible set from the observed costs:
\begin{equation}
\mathcal{A}_t(u)=\{a\in\mathcal{A}\mid c_a(u)\le B_t\}.
\end{equation}
Within this set, we favor actions that succeed, and among successful actions we favor lower cost. We implement this preference with a soft target using an accuracy--cost trade-off parameter $\alpha_{\text{pol}}\in[0,1]$:
\begin{equation}
w_a(u,t)=\mathbb{I}[a\in\mathcal{A}_t(u)]\cdot\Big(\alpha_{\text{pol}}\, y_a(u)+(1-\alpha_{\text{pol}})\,\frac{1}{c_a(u)}\Big),
\end{equation}
with $\tilde{w}_a(u,t)=w_a(u,t)/\sum_{a'}w_{a'}(u,t)$ and
\begin{equation}
\mathcal{L}_{\text{pol}}(u,t)= -\sum_{a\in\mathcal{A}}\tilde{w}_a(u,t)\log \pi_\phi(a\mid u).
\end{equation}

\noindent \textbf{Cost feasibility at inference.}
At inference, we replace $c_a(\cdot)$ with an online estimate $\hat{c}_a(v)$, for example running medians from logs adjusted by input length, and mask actions whose $\hat{c}_a(v)>B(v)$ before selecting $\arg\max_a \pi_\phi(a\mid v)$.

\subsection{Inference with Global Budget Scheduler}
\label{sec:inference_scheduler}

To enforce a user budget $B_{\text{total}}$ while still supporting multi-step reasoning, RouteGoT uses a Global Budget Scheduler that shapes graph growth based on the remaining budget.

\noindent \textbf{Synthesis reserve.}
We reserve $B_{\text{syn}}=\max(B_{\min},\alpha_{\text{syn}}B_{\text{total}})$ tokens for final synthesis, where $\alpha_{\text{syn}}$ is a fixed fraction and $B_{\min}$ is a minimum reserve. Let $C_{\text{acc}}$ denote the accumulated cost so far; the remaining expandable budget is
$B_{\text{rem}}=B_{\text{total}}-C_{\text{acc}}-B_{\text{syn}}$.
The scheduler stops expansion once $B_{\text{rem}}\le 0$.

\noindent \textbf{Complexity-conditioned depth constraints.}
We gate recursive decomposition using a node difficulty signal. A node is marked \textsf{Hard} if (i) $\hat{t}(v)=2$, or (ii) $\hat{p}_{\textsf{Decompose}}(v)$ exceeds the best non-expansion option by a margin; otherwise it is \textsf{Simple}. We allow deeper recursion only for \textsf{Hard} nodes, while \textsf{Simple} nodes are restricted to shallow execution to avoid unproductive recursion.

\noindent \textbf{Budget-regulated adaptive branching.}
When executing \textsf{Decompose}, the scheduler adapts the branching width $K(v)$ based on the remaining budget ratio $\rho=B_{\text{rem}}/B_{\text{total}}$, and penalizes width by depth and current node count so that exploration is broader early and more conservative as resources diminish.

\noindent \textbf{Plan-guided fallback mechanism.}
Even when the router selects \textsf{Decompose}, executing all generated subtasks can exceed the remaining budget. RouteGoT therefore uses an optimistic decomposition procedure: it first generates a plan (child subtasks) and then estimates the downstream execution cost. If the plan is predicted to violate the remaining budget, we trigger plan-guided fallback, reuse the plan as high-level context, and solve the current node locally via \textsf{SolveWithPlan}. This turns an over-budget expansion into a plan-and-solve step, preserving robustness while extracting value from the initial planning cost.

\section{Experiments} \label{sec:5_experiment}

We design experiments to answer the following research questions:

\noindent \textbf{RQ1:} Does \model achieve a superior accuracy--cost trade-off compared to static prompting/structured baselines and adaptive routing strategies?

\noindent \textbf{RQ2:} Can \model maintain stable performance under varying global token budgets?

\noindent \textbf{RQ3:} What are the benefits of \emph{in-process, node-level} routing compared to \emph{task-level} routing?

\noindent \textbf{RQ4:} How does \model adaptively allocate computation across nodes on real queries (qualitative analysis)?

\subsection{Experimental Setup}
\label{sec:exp_setup}

\subsubsection{Datasets}
\label{sec:datasets}

We use (i) a heterogeneous pool to train the routing components, and (ii) a set of challenging benchmarks to evaluate end-to-end performance.

\noindent \textbf{Router Training Pool.}
To train the learned components of \model, we build a heterogeneous dataset of 20{,}000 instances drawn from 12 reasoning and QA benchmarks (Table~\ref{tab:train_data_dist}). Each instance is a standalone task with a reference answer. We run three strategies, \textsf{IO}, \textsf{CoT}, and \textsf{Decompose}, and record (i) correctness $y_a(u)\in\{0,1\}$ and (ii) realized token cost $c_a(u)$ (input\,+\,output tokens) for each action $a$.
These logs supervise success prediction, budget construction via required-cost percentiles, and policy learning through budget-feasible soft targets that trade off correctness against inverse cost, with full details in Appendix~\ref{app:datasets}.

\begin{table}[t]
\centering
\captionsetup{skip=2pt}
\caption{Composition of the Router Training Dataset.}
\label{tab:train_data_dist}
\resizebox{0.95\linewidth}{!}{
\begin{tabular}{lc|lc|lc}
\toprule
\textbf{Dataset} & \textbf{Count} & \textbf{Dataset} & \textbf{Count} & \textbf{Dataset} & \textbf{Count} \\
\midrule
2WikiMultihopQA~\cite{2WikiMultihopQA} & 4,000 & WTQ~\cite{wtq} & 2,000 & GSM8K~\cite{gsm8k} & 250 \\
MuSiQue~\cite{trivedi-etal-2022-musique} & 3,250 & TabFact~\cite{chen2019tabfact} & 2,000 & SQuAD~\cite{rajpurkar-etal-2016-squad} & 250 \\
MATH~\cite{math} & 3,000 & StrategyQA~\cite{strategyqa} & 1,500 & SVAMP~\cite{svamp} & 250 \\
QASC~\cite{khot2020qasc} & 2,500 & CSQA~\cite{commonsenseqa} & 750 & OpenBookQA~\cite{openbookqa} & 250 \\
\bottomrule
\end{tabular}
}
\end{table}

\begin{table*}[t]
\centering
\scriptsize
\setlength{\tabcolsep}{2.5pt} 
\captionsetup{skip=2pt}
\caption{We report Accuracy (\%) and Average Output Tokens. All reasoning baselines utilize Qwen3-30B as the backbone. Router baselines dynamically select from a model pool comprising \{Qwen3-4B, 8B, 30B\}. The last two columns (\textbf{Avg. $\Delta$}) represent the average improvement of \model compared to the specific baseline across all seven tasks.}
\label{tab:main_all}
\resizebox{0.9\textwidth}{!}{
\begin{tabular}{l|cc|cc|cc|cc|cc|cc|cc||cc}
\toprule
\multirow{2}{*}{\textbf{Method}} & \multicolumn{2}{c|}{\textbf{GPQA}} & \multicolumn{2}{c|}{\textbf{HotpotQA}} & \multicolumn{2}{c|}{\textbf{MoreHopQA}} & \multicolumn{2}{c|}{\textbf{HybridQA}} & \multicolumn{2}{c|}{\textbf{Game of 24}} & \multicolumn{2}{c|}{\textbf{Cross.(Let)}} & \multicolumn{2}{c||}{\textbf{Cross.(Wrd)}} & \multicolumn{2}{c}{\textbf{Avg. $\Delta$ }} \\
 & Acc & Tok. & Acc & Tok. & Acc & Tok. & Acc & Tok. & Acc & Tok. & Acc & Tok. & Acc & Tok. & \textbf{Acc} & \textbf{Tok.} \\
\midrule
\multicolumn{17}{c}{\textit{Reasoning Baselines}} \\
\midrule
IO & 41.4 & 7 & 81.0 & 11 & 26.0 & 9 & 67.0 & 21 & 27.0 & 380 & 22 & 467 & 6.0 & 467 & +23.7 & +1116.2\% \\
CoT & 63.1 & 4,965 & 86.0 & 431 & 68.0 & 727 & 84.0 & 476 & 58.0 & 1,948 & 22.2 & 13,201 & \textbf{13.5} & 13,201 & +6.2 & -52.6\% \\
ToT & 44.9 & 9,077 & 60.0 & 3,748 & 61.0 & 3,860 & \textbf{91.0} & 3,538 & 20.0 & 2,855 & 12.4 & 9,563 & 2.1 & 9,563 & +20.7 & -60.8\% \\
GoT* & 59.6 & 9,468 & 87.0 & 812 & 73.0 & 814 & 88.0 & 1,635 & 72.0 & 17,396 & 22.4 & 11,674 & 6.5 & 11,674 & +3.9 & -69.0\% \\
AGoT & 64.6 & 12,179 & 72.0 & 2,583 & 70.0 & 2,064 & 84.0 & 2,097 & 74.0 & 18,406 & 11.0 & 20,893 & 3.5 & 20,893 & +8.1 & -79.1\% \\
\midrule
\multicolumn{17}{c}{\textit{Router Baselines}} \\
\midrule
Random & 60.1 & 5,658 & 77.0 & 1,685 & 64.0 & 1,516 & 70.0 & 930 & 62.0 & 7,899 & 7.6 & 6,635 & 1.0 & 6,635 & +13.5 & -46.5\% \\
KNN & 61.1 & 3,292 & 79.0 & 627 & 71.0 & 624 & 68.0 & 510 & 71.0 & 4,542 & 6.2 & 5,813 & 0.5 & 5,813 & +11.3 & -21.9\% \\
RouteLLM & 57.6 & 3,640 & 83.0 & 555 & 71.0 & 688 & 65.0 & 540 & 79.0 & 4,849 & 12.8 & 5,831 & 2.5 & 5,831 & +9.3 & -24.5\% \\
EmbedLLM & 59.6 & 11,369 & 77.0 & 2,040 & 65.0 & 2,621 & 66.0 & 1,376 & 58.0 & 9,668 & 19.0 & 6,017 & 5.0 & 6,017 & +12.4 & -57.6\% \\
RTR & 56.6 & 3,224 & 79.0 & 554 & 73.0 & 626 & 68.0 & 527 & \textbf{80.0} & 4,623 & 9.4 & 4,065 & 1.5 & 4,065 & +9.8 & -6.3\% \\
\midrule
\rowcolor{gray!10} \textbf{\model} & \textbf{65.7} & 3,352 & \textbf{88.0} & 592 & \textbf{77.0} & 665 & \textbf{91.0} & 700 & \textbf{80.0} & 3,648 & \textbf{23.4} & 3,804 & 11.0 & 3,804 & - & - \\
\bottomrule
\end{tabular}
}
\end{table*}

\noindent \textbf{Evaluation Benchmarks.}
We evaluate \model across three task categories.
For \textbf{advanced reasoning}, we use \textbf{GPQA} (Diamond split)~\cite{rein2024gpqa}, evaluating all 198 questions with shuffled options to reduce position bias.
For \textbf{retrieval \& multi-hop QA}, we use \textbf{HotpotQA}~\cite{yang-etal-2018-hotpotqa}, \textbf{MoreHopQA}~\cite{schnitzler2024morehopqa}, and \textbf{HybridQA}~\cite{chen-etal-2020-hybridqa}. To control cost, we randomly sample 100 instances from each benchmark with a fixed seed. We use the dataset-provided context (passages/tables) and do not perform external web retrieval; the exact input formatting is provided in Appendix~\ref{app:datasets}.
For \textbf{explorative reasoning}, we include \textbf{Game of 24} and \textbf{Crosswords}, following the standard test splits and evaluation protocol from the original ToT setup~\cite{tot}.

\subsubsection{Baselines}
We compare \model with two classes of baselines: standard reasoning paradigms and adaptive routing methods instantiated in our GoT-style executor.

\noindent \textbf{Standard Reasoning Paradigms.}
We include common inference methods as reference points for accuracy and cost. \textbf{IO} and \textbf{CoT}~\cite{cot} use standard zero-shot prompting and chain-of-thought prompting, respectively. For structured reasoning, we evaluate \textbf{ToT}~\cite{tot}, \textbf{GoT}~\cite{got}, and \textbf{AGoT}~\cite{agot}.
The original GoT framework assumes task-specific, hand-crafted graph schemas, which is difficult to scale to diverse, open-ended QA benchmarks. We therefore implement a generalized GoT variant that uses the same dynamic graph expansion logic as our framework, while forcing planning, execution, and synthesis to be carried out by the \textbf{Large} model. This variant avoids per-task schema design and acts as a schema-free upper bound that favors accuracy over cost.

\noindent \textbf{Adaptive Routing Baselines.}
To isolate the effect of routing, we keep the reasoning backbone and graph operations fixed and replace only the router with existing strategies. All routers use the same text embeddings and are trained (or indexed) on the same Router Training Data (Sec.~\ref{sec:datasets}). The \textbf{Random Router} samples actions uniformly; we report the mean over 10 runs. The \textbf{KNN-Router} retrieves the $k=10$ nearest training instances by embedding similarity and selects the expert with the best estimated score. For learning-based routers, we adapt three representative methods. \textbf{EmbedLLM}~\cite{zhuang2025embedllm} selects the expert with the highest predicted success probability and does not model cost. \textbf{Route-to-Reason (RTR)}~\cite{rtr} predicts both success probability and token cost and chooses the expert that maximizes a utility objective. \textbf{RouteLLM}~\cite{ong2025routellm} is a binary router for strong/weak model selection, and we adapt it to choose between our cheapest and most expensive strategies by training separate checkpoints for different cost and performance trade-offs.

\subsubsection{Implementation Details}
\label{sec:implementation}

We summarize the key configuration choices for our experimental framework. Full hyperparameters, hardware details, and baseline-specific settings are in Appendix~\ref{app:implementation}.

\noindent \textbf{Backbone Models and Strategy Binding.}
We use the \textbf{Qwen3}~\cite{yang2025qwen3technicalreport} family to instantiate actions with different capability and cost profiles. Concretely, the \textbf{Small (4B)} model runs \textsf{Direct} for simple IO prompting, the \textbf{Medium (8B)} model runs \textsf{CoT} for multi-step reasoning, and the \textbf{Large (30B)} model runs \textsf{Decompose} for planning and graph expansion. The learned components in \model use lightweight 0.6B adapters to keep routing overhead small.

\noindent \textbf{Unified Graph Environment.}
All graph-based methods, including baselines, run in the same GoT-style executor to ensure comparability. To reflect resource-limited settings, we cap the maximum graph depth at 3 and the branch at 5. We run all experiments with the vLLM inference engine on 4 NVIDIA RTX A6000 GPUs.

\noindent \textbf{Router and Baseline Setup.}
All learning-based routers share the same sentence encoder for node inputs, using \texttt{all-mpnet-base-v2}\footnote{https://huggingface.co/sentence-transformers/all-mpnet-base-v2}. For cost-aware baselines (KNN, RTR, and RouteLLM), we report three operating points (Performance, Balanced, and Cost-Saving) obtained by varying the cost penalty coefficient $\beta$. Unless stated otherwise, we use the \textit{Balanced} setting in the main comparisons.

\subsubsection{Evaluation Metrics}
\label{sec:metrics}

For multiple-choice GPQA, we report exact-match accuracy on the chosen option after shuffling the answer choices.
For open-ended QA and reasoning benchmarks, we evaluate semantic equivalence with a Qwen3-30B judge: given a prediction $\hat{y}$ and reference $y$, the judge returns a binary correctness label (see Appendix~\ref{app:prompts} for the prompt). The judge is blind to method identity and uses deterministic decoding. For explorative reasoning tasks, we follow the evaluation protocol from ToT~\cite{tot}. We report average input and output tokens per query and ouput tokens are computed as the sum over all model calls during reasoning, including planning, node execution, synthesis, and router overhead when applicable. Tokens consumed by the evaluation judge are excluded from system cost because the judge is only used offline.

\begin{table}[t]
\centering
\captionsetup{skip=2pt}
\caption{\textbf{Per Question Inference Time Comparison (Seconds).}}
\label{tab:inference_time}
\resizebox{0.8\linewidth}{!}{
\begin{tabular}{l|ccc|c}
\toprule
\textbf{Dataset} & \textbf{GoT*} & \textbf{AGoT} & \textbf{RTR} & \textbf{RouteGoT} \\
\midrule
GPQA & 97.01 & 42.69 & \textbf{31.70} & 32.64 \\
Game of 24 & 98.20 & 112.63 & 62.13 & \textbf{60.93} \\
Crosswords & 116.03 & 119.00 & 47.00 & \textbf{37.09} \\
HybridQA & 38.66 & 42.83 & \textbf{10.80} & 12.93 \\
MoreHopQA & 70.38 & 17.72 & \textbf{7.27} & 11.79 \\
HotpotQA & 22.89 & 23.93 & \textbf{5.46} & 6.95 \\
\bottomrule
\end{tabular}
}
\end{table}

\subsection{Performance and Efficiency Analysis (RQ1)}
\label{sec:main_results}

Table~\ref{tab:main_all} and Table~\ref{tab:inference_time} present the performance of \model across seven benchmarks. For brevity, we report average output tokens in Table~\ref{tab:main_all}, with input tokens provided in Table~\ref{tab:appendix_input_tokens}. 

\model exhibits a favorable balance between reasoning accuracy and computational efficiency, reaching or sharing the top performance across all evaluated tasks. Compared to structured baselines such as AGoT, \model provides an average accuracy gain of 8.1 percentage points while reducing output token consumption by 79.1\%. The high cost associated with AGoT is primarily attributed to repeated graph executions involving extensive context concatenation and lengthy prompts, leading to substantial input token overhead on certain datasets. These efficiency gains remain consistent across diverse tasks; for example, on HotpotQA, accuracy increases from 72.0\% to 88.0\% alongside a 77.1\% reduction in output tokens. This decrease in token volume corresponds to lower inference latency, as \model consistently outperforms AGoT in speed and achieves nearly a $6\times$ speedup over GoT* on MoreHopQA. Adaptive routing baselines like RTR often sacrifice solution quality for speed, whereas \model maintains high accuracy on complex reasoning tasks. While RTR achieves low latency, \model exceeds its performance by an average of 9.8 points with comparable token usage. On HybridQA, \model improves accuracy to 91.0\% from the 68.0\% achieved by RTR. Furthermore, on search-intensive tasks such as Crosswords, \model is both more accurate and faster than RTR, requiring 37.09s compared to 47.00s. These results suggest that rather than pruning indiscriminately to meet budget constraints, \model strategically allocates computational resources to critical nodes, thereby avoiding the performance degradation typical of aggressive cost-reduction strategies.

\begin{figure}[t]
  \centering
  \includegraphics[width=0.9\linewidth]{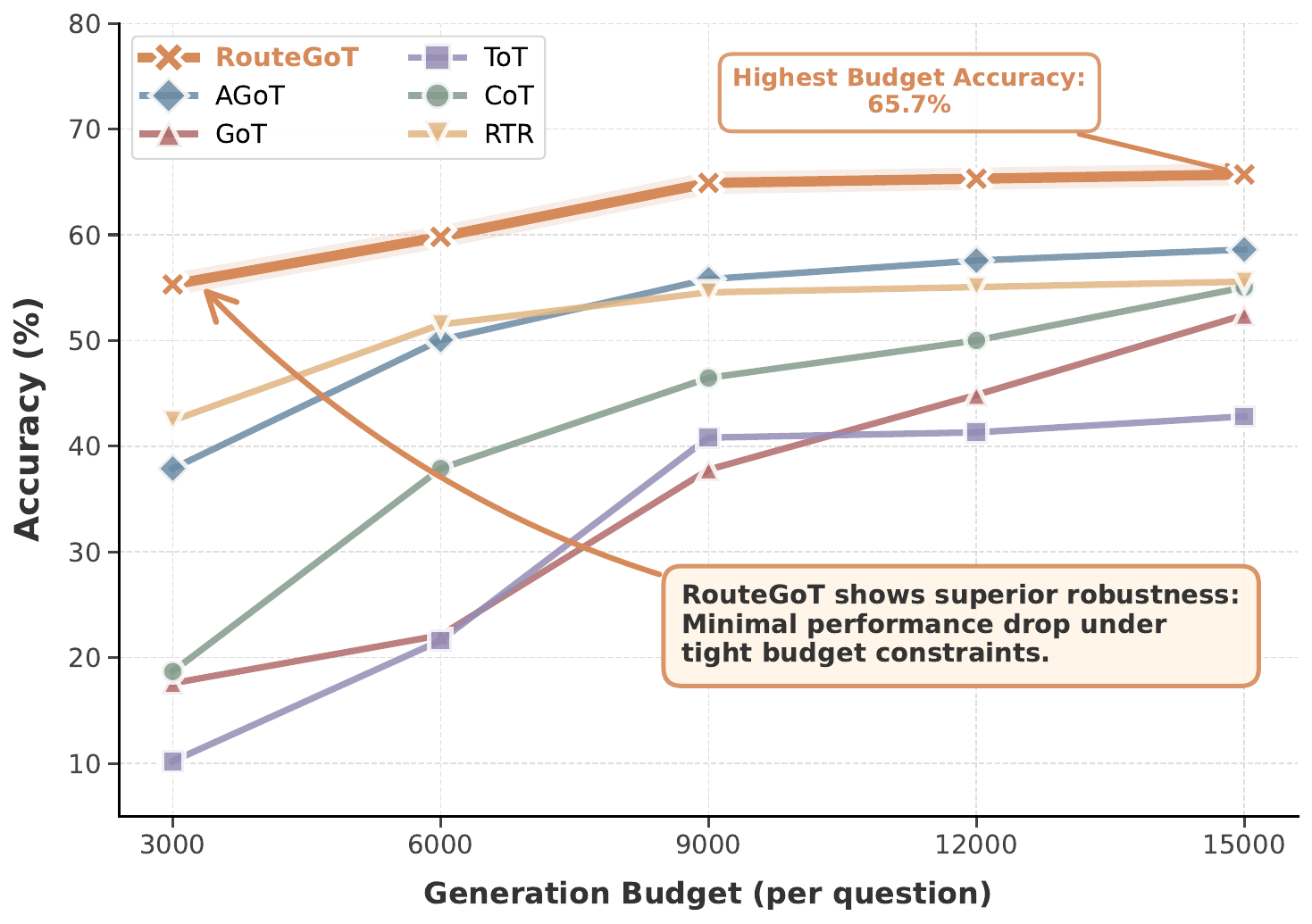}
  \captionsetup{skip=2pt}
  \caption{Performance comparison under varying computational budgets on GPQA.}
  \label{fig:cost_accuracy_tradeoff}
\end{figure}

\begin{figure*}[t]
  \centering
  \begin{minipage}{0.32\textwidth}
    \centering
    \includegraphics[width=\linewidth]{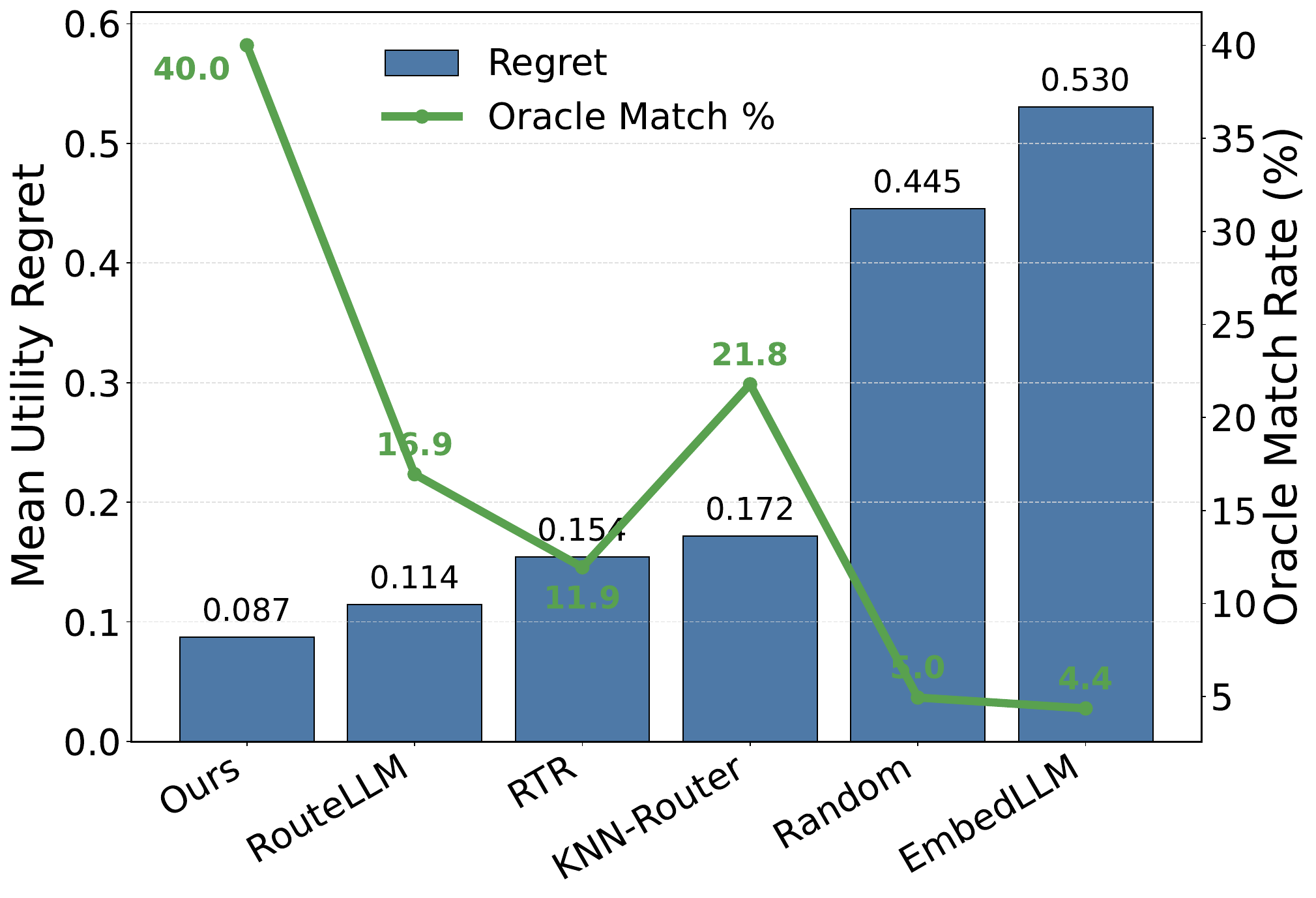}
    \centerline{\small (a) Decision Quality (Lower regret is better)}
  \end{minipage}
  \hfill
  \begin{minipage}{0.32\textwidth}
    \centering
    \includegraphics[width=\linewidth]{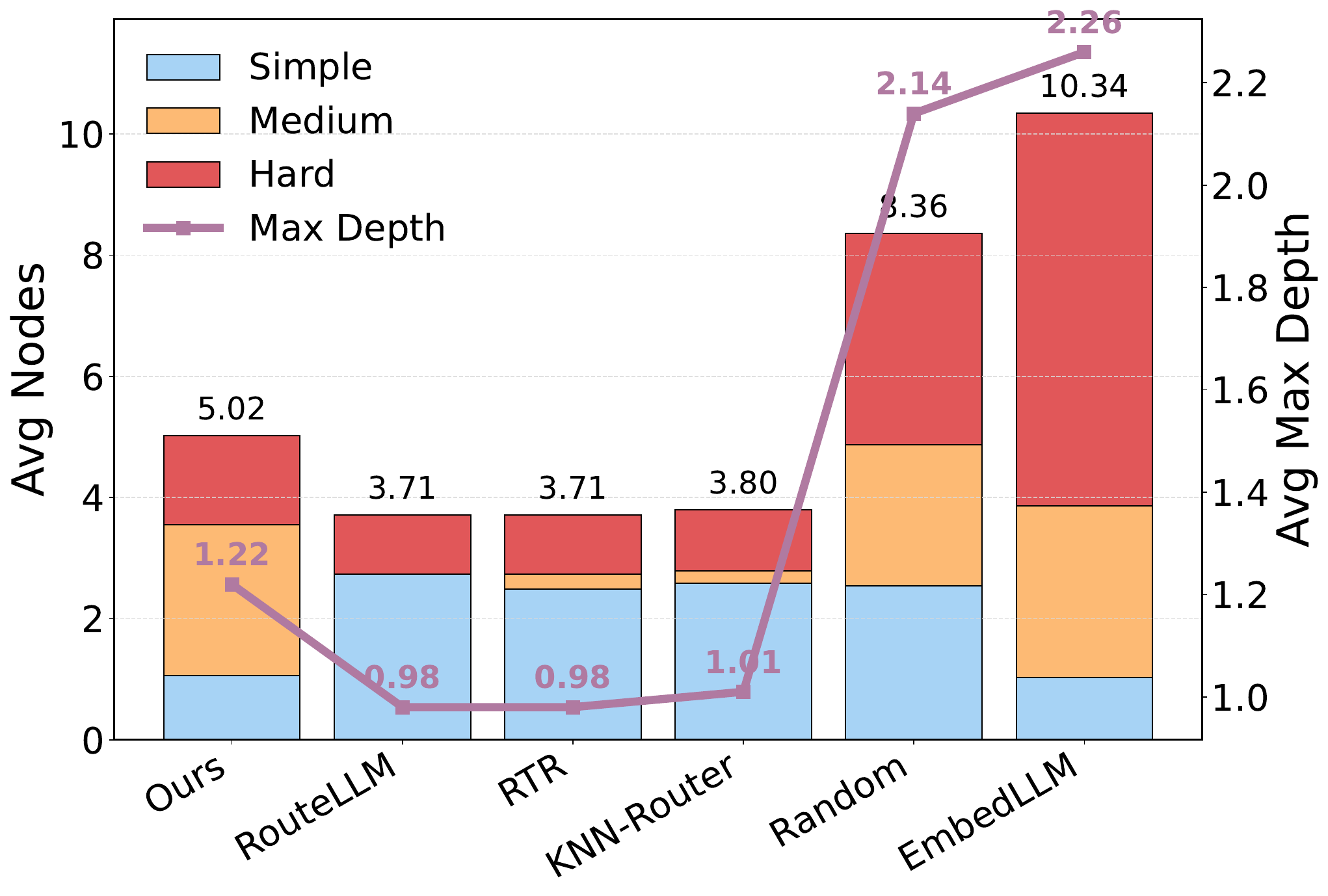} 
    \centerline{\small (b) Node Distribution}
  \end{minipage}
  \hfill
  \begin{minipage}{0.32\textwidth}
    \centering
    \includegraphics[width=\linewidth]{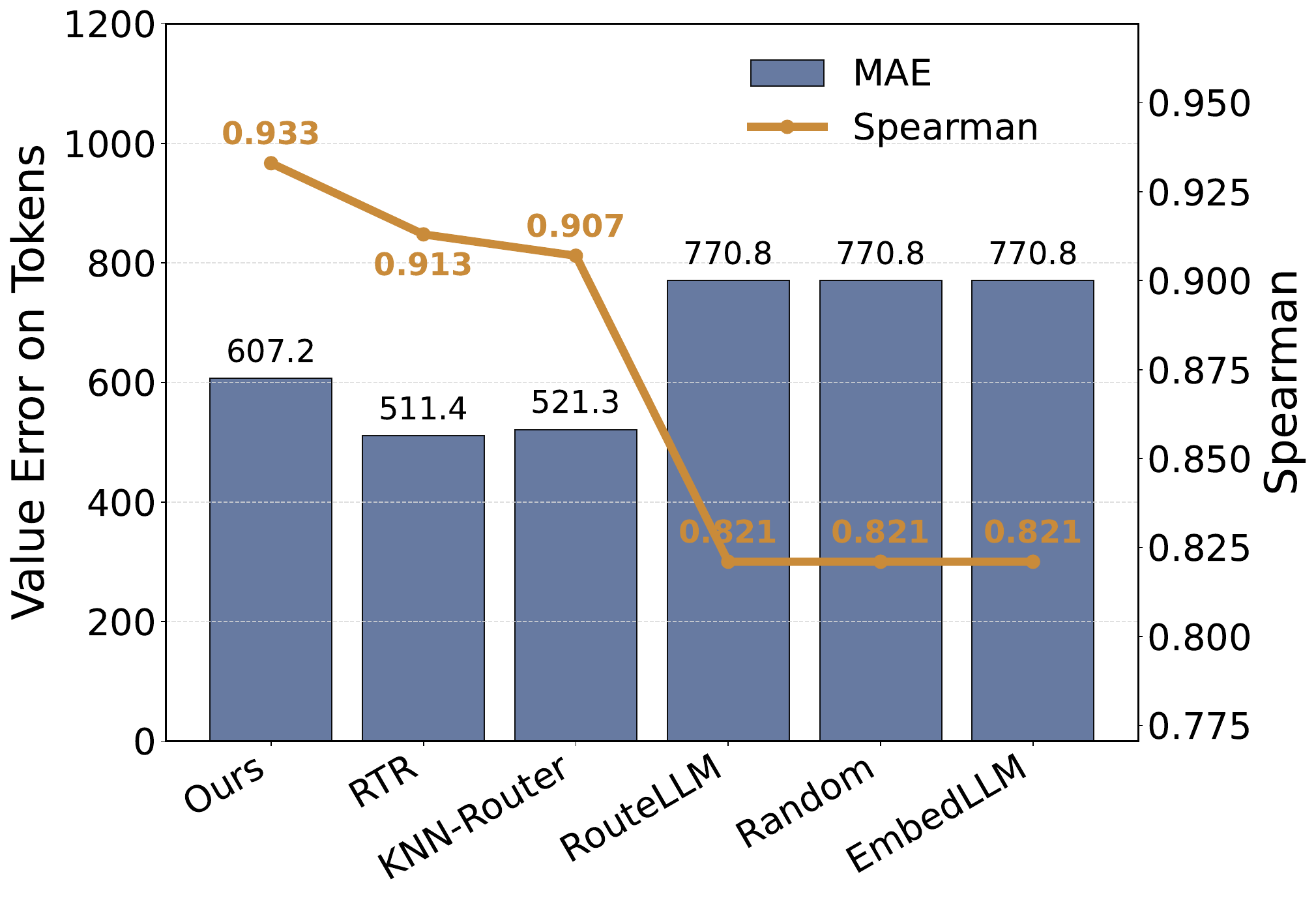} 
    \centerline{\small (c) Cost Analysis}
  \end{minipage}
  
  \vspace{-0.2cm}
  
  \caption{\textbf{Evaluation of Routing Mechanism (RQ3).} 
  \textbf{(a)} Comparison of decision quality metrics (Regret, Utility). 
  \textbf{(b)} Distribution of node complexity across different methods.
  \textbf{(c)} Performance analysis of the cost predictor.}
  \label{fig:router_metrics} 
\end{figure*}

\subsection{Budget Controllability and Stability (RQ2)}
\label{sec:budget_analysis}

Figure~\ref{fig:cost_accuracy_tradeoff} shows how accuracy changes as we vary the per-question compute budget. We sweep a global token cap $B_{\text{total}}$, enforced on total tokens across all model calls.

\model remains reliable under tight constraints. At the lowest budget, it reaches 54.3\% accuracy, exceeding AGoT by 16.4 points and RTR by 10.9 points. This pattern suggests that in-graph routing can still prioritize the most useful expansions when budget limits sharply restrict search. Several structured baselines show a clear cold-start behavior: under small budgets, ToT, GoT*, and CoT fall below the 4-way random-guessing level (25\%) and only become competitive after the budget increases. As $B_{\text{total}}$ grows, \model improves steadily, tracks the best curve across the sweep, and peaks at 65.7\%. Its performance declines gradually as budgets tighten, with an 11.4-point gap between the strictest and most lenient settings. By comparison, AGoT is more budget-sensitive, varying by about 20 points, and CoT, GoT*, and ToT need substantially larger budgets to approach their best operating regions. Even at the highest budget, \model keeps an advantage of about 7 points over AGoT. Taken together, the results indicate that budget-aware scheduling and node-level routing support both low-budget robustness and a higher accuracy ceiling under practical constraints.

A practical factor is that graph-based methods pay noticeable input overhead from repeated context injection and multi-call prompts. When $B_{\text{total}}$ limits total tokens, this overhead tightens the effective budget more than output-token counts alone would imply, especially for methods with heavier prompting such as AGoT. \model reduces this cost by using compact branch summaries and lightweight routing, leaving more of the budget for the reasoning steps that most affect the final prediction.

\subsection{Analysis of Routing Mechanism (RQ3)}
\label{sec:router_analysis}

To isolate the effect of routing, we compare routers within a unified graph executor on HotpotQA, fixing the global budget to $B_{\text{total}}=20{,}000$ total tokens and the maximum depth to 3. Under this controlled setup, differences in outcomes mainly capture how each router allocates compute and shapes the reasoning topology, rather than artifacts of the execution backend.

We measure decision quality using Utility Regret and Oracle Match Rate (Figure~\ref{fig:router_metrics}a). For a question $i$ and method $m$, we define utility as $U(i,m)=\mathrm{Acc}(i,m)-\beta\cdot \mathrm{Cost}(i,m)$, where $\beta$ is the balanced operating point used by our cost-aware baselines. Regret is $\mathrm{Regret}(i,m)=\max_{m'\in M}U(i,m')-U(i,m)$ over the comparison set $M$, and the Oracle Match Rate is the fraction of questions on which a method attains the highest utility. \model achieves the lowest mean regret (0.087) and the highest oracle match rate (40.0\%). This rate is nearly twice that of the strongest baseline, KNN (21.8\%), suggesting that \model more consistently chooses high-utility reasoning paths under budget constraints. These trends are consistent with the accuracy--cost trade-offs in Table~\ref{tab:main_all}.

Figure~\ref{fig:router_metrics}(b) illustrates the induced node distribution and effective depth. Cost-biased routers such as RouteLLM and RTR tend to produce very shallow graphs (averaging 3.71 nodes and 0.98 max depth), often terminating reasoning prematurely and missing necessary decompositions. Conversely, EmbedLLM and Random approaches tend to over-expand, wasting budget on redundant subproblems. \model strikes a balance, with an average of 5.02 nodes and a max depth of 1.22. This supports the hypothesis that the system is ``deep enough but lean enough,'' expanding selectively only when the expected utility warrants it.

Finally, Figure~\ref{fig:router_metrics}(c) evaluates the quality of token-cost estimation. We observe that exact token prediction remains noisy even for learned methods, with Mean Absolute Error (MAE) ranging from 511 to 607 tokens for RTR, KNN, and \model. Routers lacking learned cost signals rely on static proxies, incurring significantly larger errors. However, ranking costs is more feasible than exact prediction. \model achieves the highest Spearman correlation, outperforming RTR and KNN, while proxy-based methods lag at 0.821. This is consistent with our objective: the Budget Predictor is trained for ordinal feasibility and penalizes under-estimation more heavily, so rank correlation and under-estimation risk align better with hard-budget routing than MAE.

\begin{figure*}[t]
  \centering
  \includegraphics[width=0.9\textwidth]{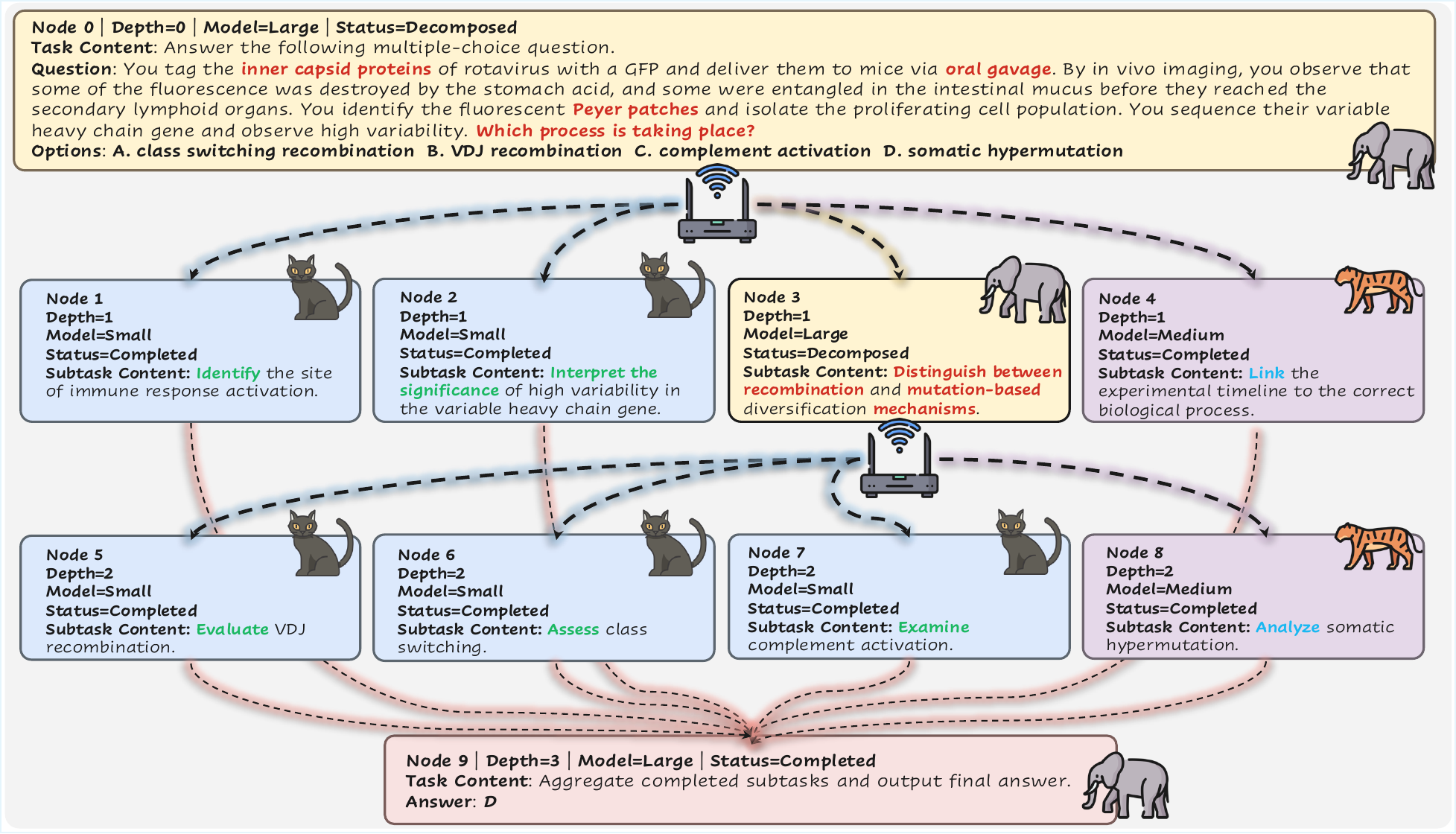}
  \captionsetup{skip=1pt}
\caption{Case study on a GPQA biomedical task. \model\ dynamically routes nodes based on semantic triggers. Red nodes represent \textsf{Decompose} actions for high-level planning, blue nodes indicate \textsf{CoT} for relational analysis, and green nodes denote \textsf{IO} for efficient pruning. Compared to AGoT, \model\ avoids redundant sub-graph expansion on distractors, identifying the correct answer with substantially reduced token consumption.}
  \label{fig:case_study}
\end{figure*}

\subsection{Qualitative Case Study (RQ4)}
\label{sec:case_study}

Figure~\ref{fig:case_study} demonstrates the efficiency of \model on a complex biomedical query from GPQA. After the initial decomposition, the router detects a spatial inconsistency in the VDJ Recombination distractor, noting the conflict between its occurrence in bone marrow and the experimental observation in Peyer patches. Rather than generating a computationally expensive sub-tree to explain molecular mechanisms, \model employs a small model to prune this branch through IO. This resource-saving logic extends to other distractors, allowing computational efforts to be concentrated on the correct path: Somatic Hypermutation. Such selective expansion preserves reasoning depth where required while eliminating redundancy. In this instance, the AGoT baseline exhibits blind expansion, consuming 16,637 tokens only to reach an incorrect conclusion. In contrast, \model identifies the correct result using only 2,622 tokens, representing a 6.3$\times$ improvement in cost-efficiency.

\subsection{Ablation Study}
\label{sec:ablation}

We study the role of each learned component in \model with ablations on HotpotQA, MoreHopQA, and HybridQA. We compare the full system with two variants. The \textbf{w/o Budget Predictor (BP)} variant replaces the ordinal Budget Predictor with a continuous cost regressor, testing whether coarse, discrete budgets provide a more reliable signal than direct cost regression. The \textbf{w/o BP + PolicyNet (PN)} variant further removes PolicyNet and selects actions using Success Predictor scores, which tests whether learned decision rules are necessary beyond confidence-based routing.

\begin{table}[htbp]
\centering
\captionsetup{skip=2pt}
\caption{Comparison of \model against variants with ablated components.}
\label{tab:ablation}
\resizebox{0.95\linewidth}{!}{
\begin{tabular}{l|cc|cc|cc}
\toprule
\multirow{2}{*}{\textbf{Method}} & \multicolumn{2}{c|}{\textbf{HotpotQA}} & \multicolumn{2}{c|}{\textbf{MoreHopQA}} & \multicolumn{2}{c}{\textbf{HybridQA}} \\
& \textbf{Acc (\%)} & \textbf{Tokens} & \textbf{Acc (\%)} & \textbf{Tokens} & \textbf{Acc (\%)} & \textbf{Tokens} \\
\midrule
\rowcolor{gray!10} \textbf{\model} & 88.0 & 592 & 77.0 & 665 & 91.0 & 700 \\
\midrule
w/o BP & 86.0 & 2438 & 78.0 & 2649 & 89.0 & 1634 \\
w/o BP + PN & 78.0 & 2020 & 70.0 & 2951 & 83.0 & 1722 \\
\bottomrule
\end{tabular}
}
\end{table}

\noindent \textbf{Analysis.}
Table~\ref{tab:ablation} reports the results. Replacing the Tier Predictor with a continuous regressor (w/o BP) consistently increases token use, nearly quadrupling cost on HotpotQA (592 vs. 2438) and similarly inflating MoreHopQA. Accuracy stays close, but largely because the router spends more tokens, so efficiency drops sharply. This supports the use of ordinal tiers as a stable mechanism for enforcing per-node budgets, rather than relying on noisy point estimates that drift toward expensive choices. Removing PolicyNet as well (w/o BP+PN) reduces accuracy substantially, with HotpotQA falling from 88.0\% to 78.0\%. This suggests that confidence scores alone are not enough to choose effective graph actions, especially when deciding whether to expand a subgraph. PolicyNet provides the missing coupling between success signals, budget constraints, and action selection that improves the overall cost accuracy balance.

\section{Related Work} \label{sec:6_related}

\subsection{LLM Reasoning}
Existing reasoning methods can be broadly grouped into linear prompting and structured reasoning with explicit topologies~\cite{zhou2023leasttomostpromptingenablescomplex,anonymous2026dagmath,zhang2026diagramthought,wei2025graphchain,li2025graphif,huan2025scaling}. Linear prompting is represented by CoT~\cite{cot}, which improves the solvability of complex problems by explicitly generating intermediate steps. To mitigate the instability of single-path reasoning and error propagation, Self-Consistency CoT (SC-CoT)~\cite{sc-cot} and related approaches sample multiple reasoning trajectories for the same input and aggregate the final answers to improve robustness. Structured reasoning further introduces explicit topologies to expand the search and composition space: ToT~\cite{tot} organizes reasoning as a tree and performs search to select better paths; GoT~\cite{got} uses a more general graph structure to represent dependencies and enables more flexible composition and information propagation; subsequent work such as AGoT~\cite{agot} introduces adaptive expansion within the GoT framework and can yield strong performance on challenging reasoning benchmarks. However, these gains often come with substantial inference overhead, since adaptive expansion still relies on heavyweight model calls for planning, node assessment, and recursive decomposition, making end-to-end cost and latency difficult to control. To our knowledge, prior work has not provided an efficient AGoT-style inference scheme that performs fine-grained, in-process compute allocation across graph nodes under explicit budget constraints. In contrast, we focus on efficient GoT/AGoT execution by routing each node to a model-bound reasoning action based on predicted difficulty and remaining budget, and by scheduling expansion and execution within a global budget to reduce redundant computation while preserving solution quality.

\subsection{Routing and Model/Strategy Selection}
Routing methods can be summarized along two axes: decision granularity and decision targets~\cite{chen2025harnessing,chen2025putting,hao2025llmtmbenchmarkingoptimizingllms,feng2025graphrouter,anonymous2026radar}. For task-level routing, cascading or multi-call methods such as FrugalGPT~\cite{chen2024frugalgpt}, LLM-Blender~\cite{llmblender}, and AutoMix~\cite{aggarwal2024automix} adopt weak-to-strong escalation or multi-model ensembling and fusion to improve quality, but they typically increase the number of calls and end-to-end latency. Supervised one-shot routing methods such as Hybrid-LLM~\cite{hybridllm} and RouteLLM~\cite{ong2025routellm} train lightweight routers to decide, based on the query, whether to escalate to a stronger model, aiming to reduce cost while maintaining quality and generalization. Beyond model-only routing, joint routing over model $\times$ reasoning strategy, represented by Route-To-Reason (RTR)~\cite{rtr}, considers both success likelihood and token cost and selects the best combination via a trade-off coefficient. Other lines of work study more general routing for new models and distribution shifts, such as ZeroRouter~\cite{yan2026breaking}, as well as subtask-level collaboration with reinforcement learning, such as Route-and-Reason~\cite{rnr}. Different from these approaches, our routing happens inside the graph reasoning process: instead of making a single decision for the entire task, we dynamically allocate models and strategies for subtask nodes at the step/node level, coupled with budget control.

\section{Conclusion}
\label{sec:conclusion}

In this work, we introduced \model to reduce the cost of static graph-structured reasoning. Instead of committing to a single strategy for an entire problem, \model makes routing decisions at the node level, using larger models only for planning and synthesis when they are most likely to matter. Across seven benchmarks, our approach improves the accuracy--cost trade-off and outperforms both fixed graph baselines and heuristic routers. A promising direction is to place \model within a hierarchical routing pipeline, where a lightweight gateway estimates query difficulty and skips graph construction for straightforward inputs, invoking node-level graph routing only when structured reasoning is warranted. Overall, our results indicate that budget-aware control at fine granularity is an effective way to scale complex reasoning in practice.

\bibliographystyle{ACM-Reference-Format}
\bibliography{reference}

\newpage
\appendix
\section{Algorithm Details}
\label{sec:appendix_algorithm}

In this section, we provide the detailed inference procedure of \model. As discussed in Section~\ref{sec:inference_scheduler}, the framework balances reasoning depth and computational cost through a combination of node-level routing and a global budget scheduler.

\subsection{Full Inference Procedure}
Algorithm~\ref{alg:routegot_inference} outlines the step-by-step execution of \model. The process begins with a root decomposition and proceeds with a budget-aware loop. 

\begin{algorithm}[htbp]
\caption{\model Inference with Global Budget Scheduling}
\label{alg:routegot_inference}
\SetAlgoLined
\DontPrintSemicolon
\KwIn{Initial question $x$, total budget $B_{\text{tot}}$, reserve fraction $\alpha$, minimum synthesis buffer $B_{\min}$}
\KwOut{Final synthesized answer $\hat{y}$}

\tcp{Initialization and Budget Allocation}
$C_{\text{acc}} \gets 0$ \tcp*[r]{Accumulated token cost}
$B_{\text{syn}} \gets \max(B_{\min}, \alpha B_{\text{tot}})$ \tcp*[r]{Reserve budget for final synthesis}
$B_{\text{rem}} \gets B_{\text{tot}} - B_{\text{syn}}$ \tcp*[r]{Available budget for graph expansion}
$\mathcal{Q} \gets \emptyset$ \tcp*[r]{Priority queue for active reasoning nodes}

\tcp{Root Decomposition}
$(\{v_i\}_{i=1}^n, s_0) \gets \textsf{Decompose}(x)$ \tcp*[r]{Initial planning step}
$C_{\text{acc}} \gets C_{\text{acc}} + \text{Cost}_{\text{dec}}$\;
$\text{Enqueue}(\mathcal{Q}, \{v_i\})$ with initial context $s_0$\;

\While{$\mathcal{Q} \neq \emptyset$}{
    \tcp{Check Termination Condition}
    \If{$B_{\text{rem}} \le 0$}{
        \textbf{break} \tcp*[r]{Exit loop if budget is exhausted}
    }
    
    $v \gets \text{pop}(\mathcal{Q})$\;
    $(\hat{p}, \hat{t}) \gets \textsf{PredictDifficulty}(v)$ \tcp*[r]{Predict success prob. and compute budget}
    $B_v \gets \min(B_{\text{rem}}, B_{\text{cap}}(\hat{t}))$ \tcp*[r]{Cap node budget by predicted tier}
    
    \tcp{Action Selection under Constraints}
    $\hat{a} \gets \arg\max_{a \in \mathcal{A}} \pi_\phi(a \mid v) \text{ s.t. } c_a \le B_v$\;

    \uIf{$\hat{a} = \textsf{Decompose}$ \textbf{\textup{and}} $\textsf{CheckDepth}(v)$}{
        \tcp{Explorative Expansion or Plan-guided Fallback}
        $(\{v_j\}, s') \gets \textsf{Decompose}(v)$\;
        $C_{\text{acc}} \gets C_{\text{acc}} + \text{Cost}_{\text{dec}}$\;
        
        \uIf{$C_{\text{acc}} + \textsf{EstimateSubtree}(\{v_j\}) \le B_{\text{rem}}$}{
            $\text{Enqueue}(\mathcal{Q}, \{v_j\})$ with plan $s'$ \tcp*[r]{Recursive expansion}
        }
        \Else{
            $r(v) \gets \textsf{SolveWithPlan}(v, \{v_j\})$ \tcp*[r]{Budget-constrained fallback}
            $C_{\text{acc}} \gets C_{\text{acc}} + \text{Cost}_{\text{fb}}$\;
        }
    }
    \Else{
        \tcp{Atomic Execution: IO or CoT}
        $r(v) \gets \textsf{Execute}(\hat{a}, v)$\;
        $C_{\text{acc}} \gets C_{\text{acc}} + \text{Cost}_{\text{exec}}$\;
    }
    $B_{\text{rem}} \gets B_{\text{tot}} - C_{\text{acc}} - B_{\text{syn}}$ \tcp*[r]{Refresh remaining budget}
}

\tcp{Global Convergence}
\Return $\textsf{Synthesis}(x, \{r(v)\})$ \tcp*[r]{Consolidate all node results into $\hat{y}$}
\end{algorithm}

\subsection{Sub-routine Definitions}
To provide further clarity, we define the sub-routines utilized in Algorithm~\ref{alg:routegot_inference}. The \textsf{CheckDepth}($v$) function serves as a safety guard that returns true only if the current node's depth is within the maximum allowed threshold and the node is classified as \textsf{Hard} according to the criteria in Section~\ref{sec:inference_scheduler}. 

Furthermore, \textsf{EstCost}($\{v_j\}$) estimates the potential future cost of a newly generated plan by calculating the expected consumption of child nodes based on their predicted compute tiers. Finally, the \textsf{SolveWithPlan}($v, \{v_j\}$) routine implements the fallback mechanism discussed in Section~\ref{sec:inference_scheduler}. This procedure leverages the generated child subtasks $\{v_j\}$ as supplemental context for a local solver, enabling the system to address node $v$ without incurring further branching costs when budget constraints are tight.

\section{Dataset Details and Benchmark Specifications}
\label{app:datasets}

In this section, we provide extended details regarding the training data used for the routing components and the specific protocols for end-to-end evaluation.

\subsection{Router Training Pool Configuration}
The training pool consists of 20,000 instances sampled from 12 reasoning benchmarks to ensure that the learned router generalizes across various patterns, including mathematical logic and multi-hop retrieval. For offline data collection, we maintain consistency with our evaluation by employing Qwen3-4B as the small model for \textsf{Direct} actions, Qwen3-8B as the medium model for \textsf{CoT}, and Qwen3-30B as the large model for the \textsf{Decompose} action. The compute tiers are derived from the empirical distribution of the required cost $c_{\text{req}}(u)$, defined as the minimum cost among all successful actions. Based on the training logs, the 25th percentile ($B_{25}$) is 185 tokens and the 75th percentile ($B_{75}$) is 1,186 tokens. These define our three-tier difficulty classification used by the PolicyNet to mask actions exceeding the predicted tier's budget cap.

\subsection{Evaluation Benchmarks and Protocols}
We employ a fixed random seed of 42 for all sampling and shuffling operations during evaluation. For each dataset, we apply a global shuffle before selecting the instances: the full 198 questions of the GPQA Diamond split, a range of 901--1000 from the Game of 24 \textit{ToT} split, 20 games for MiniCrossword, and 100 randomly sampled instances for HotpotQA, MoreHopQA, and HybridQA. Specifically, for these sampled datasets, we prioritize extracting instances from the official test set; if a test set is unavailable, we sample from the validation set, and in cases where neither is explicitly provided, we sample directly from the available dataset pool. The input formatting for retrieval tasks parses the context into a concatenated text format of titles followed by sentences, while HybridQA utilized AGoT-style serialization for tabular data to preserve relational dependencies.

Correctness is determined through specialized protocols for each category. For GPQA, we require a strict match of the predicted option letter. For retrieval-based QA tasks, we utilize Qwen3-30B as a judge to evaluate semantic consistency with the reference. Game of 24 success is adjudicated through a rule-based validator verifying expression parsing, digit multiset consistency, and a final result of exactly 24. For MiniCrossword, the primary metric is the exact match of the $5 \times 5$ grid, supplemented by word-level and letter-level accuracy.

\section{Implementation Details}\label{app:implementation}
This section describes the technical specifications for the hardware environment, model architectures, and optimization protocols used in the \model\ framework.

\subsection{Hardware Infrastructure and Serving}Experiments were conducted on a GPU cluster equipped with four NVIDIA RTX A6000 (48GB) units. The Qwen3 model family was deployed using the vLLM serving engine to support multi-tier reasoning. To optimize memory usage and throughput, the 30B model was served across two GPUs via tensor parallelism, while the 8B and 4B models were each assigned to a single GPU. This configuration ensures that all heterogeneous experts remain resident in memory, allowing for a realistic assessment of latency and cost-accuracy trade-offs.

\subsection{Architectures of Learned Components}The routing framework consists of three neural components. The Success Predictor is based on a Qwen3-4B-Instruct model configured for sequence classification with three output heads, which provide logit-based scores for the \textsf{Direct}, \textsf{CoT}, and \textsf{Decompose} strategies. The Budget Predictor uses a Qwen3-0.6B backbone in an ordinal regression configuration with two binary heads to predict the probabilities $P(\text{Budget} \ge 1)$ and $P(\text{Budget} \ge 2)$. The PolicyNet is implemented as a three-layer Multi-Layer Perceptron (MLP) with a hidden dimension of 32 and ReLU activation. The input vector $\mathbf{z}(u) \in \mathbb{R}^6$ concatenates the three-dimensional logits from the Success Predictor with a three-dimensional one-hot encoding of the predicted tier. This design bypasses high-dimensional sentence embeddings to minimize routing overhead.

\subsection{Training Protocols and Hyperparameters}All learned components were optimized using AdamW. The Success Predictor was trained for 5 epochs with a learning rate of $2 \times 10^{-5}$, weight decay of 0.05, and a warmup ratio of 0.1. The Tier Predictor was similarly trained for 5 epochs ($LR = 2 \times 10^{-5}$) using an asymmetric cost matrix ($\alpha_{\text{asym}} = 0.1$) with specific penalties for under-estimation ($c_{20} = 3.0$, $c_{21} = 1.5$) to ensure sufficient reasoning depth. The corresponding inference thresholds were set to $\tau_1 = 0.1$ and $\tau_2 = 0.5$. The PolicyNet was trained for 10 epochs ($LR = 3 \times 10^{-4}$, weight decay $1 \times 10^{-4}$) using a rank-based objective. For tier construction, empirical thresholds $B_{25} = 185.0$ and $B_{75} = 1,185.5$ were defined based on the required-cost distribution in the training pool.

\subsection{Online Execution and Baselines}To ensure stability during inference, \model\ uses fixed cost estimates for budget-aware scheduling. For the RouteLLM baseline, we configured binary selection logic between \textsf{IO} (Weak) and \textsf{Decompose} (Strong) actions with a decision threshold of 0.5. All graph-based methods are subject to a global node limit of 50 to prevent unbounded recursion.

\begin{table*}[t]
\centering
\scriptsize
\setlength{\tabcolsep}{3pt} 
\captionsetup{skip=2pt}
\caption{Detailed evaluation results reporting Accuracy (\%) and \textbf{Average Input Tokens}. All reasoning baselines utilize Qwen3-30B. Router baselines select from the \{Qwen3-4B, 8B, 30B\} pool. This table complements the main results by highlighting the context processing overhead across different tasks.}
\label{tab:appendix_input_tokens}
\resizebox{\textwidth}{!}{
\begin{tabular}{l|cc|cc|cc|cc|cc|cc|cc}
\toprule
\multirow{2}{*}{\textbf{Method}} & \multicolumn{2}{c|}{\textbf{GPQA}} & \multicolumn{2}{c|}{\textbf{HotpotQA}} & \multicolumn{2}{c|}{\textbf{MoreHopQA}} & \multicolumn{2}{c|}{\textbf{HybridQA}} & \multicolumn{2}{c|}{\textbf{Game of 24}} & \multicolumn{2}{c|}{\textbf{Cross.(Let)}} & \multicolumn{2}{c}{\textbf{Cross.(Wrd)}} \\
 & Acc & In.Tok & Acc & In.Tok & Acc & In.Tok & Acc & In.Tok & Acc & In.Tok & Acc & In.Tok & Acc & In.Tok \\
\midrule
\multicolumn{15}{c}{\textit{Reasoning Baselines}} \\
\midrule
IO & 41.4 & 250 & 81.0 & 1,685 & 26.0 & 442 & 67.0 & 10,462 & 27.0 & 125 & 22.0 & 179 & 6.0 & 179 \\
CoT & 63.1 & 247 & 86.0 & 1,682 & 68.0 & 439 & 84.0 & 10,239 & 58.0 & 123 & 22.2 & 182 & 13.5 & 182 \\
ToT & 44.9 & 4,928 & 60.0 & 2,416 & 61.0 & 2,027 & 91.0 & 12,355 & 20.0 & 23,334 & 12.4 & 790 & 2.1 & 790 \\
GoT* & 59.6 & 16,826 & 87.0 & 10,864 & 73.0 & 4,275 & 88.0 & 65,134 & 72.0 & 19,814 & 22.4 & 13,597 & 6.5 & 13,597 \\
AGoT & 64.6 & 35,564 & 72.0 & 25,269 & 70.0 & 9,863 & 84.0 & 112,953 & 74.0 & 28,436 & 11.0 & 9,522 & 3.5 & 9,522 \\
\midrule
\multicolumn{15}{c}{\textit{Router Baselines}} \\
\midrule
Random & 60.1 & 7,823 & 77.0 & 6,656 & 64.0 & 4,512 & 70.0 & 21,138 & 62.0 & 10,909 & 7.6 & 8,767 & 1.0 & 8,767 \\
KNN & 61.1 & 4,039 & 79.0 & 4,604 & 71.0 & 2,494 & 68.0 & 20,686 & 71.0 & 5,801 & 6.2 & 5,044 & 0.5 & 5,044 \\
RouteLLM & 57.6 & 4,358 & 83.0 & 4,479 & 71.0 & 2,534 & 65.0 & 20,652 & 79.0 & 6,104 & 12.8 & 5,110 & 2.5 & 5,110 \\
EmbedLLM & 59.6 & 15,387 & 77.0 & 7,405 & 65.0 & 6,842 & 66.0 & 21,875 & 58.0 & 13,339 & 19.0 & 4,846 & 5.0 & 4,846 \\
RTR & 56.6 & 3,998 & 79.0 & 4,480 & 73.0 & 2,468 & 68.0 & 20,839 & 80.0 & 5,870 & 9.4 & 3,777 & 1.5 & 3,777 \\
\midrule
\rowcolor{gray!10} \textbf{\model} & 65.7 & 3,696 & 88.0 & 4,566 & 77.0 & 3,396 & 91.0 & 14,743 & 80.0 & 4,484 & 23.4 & 4,222 & 11.0 & 4,222 \\
\bottomrule
\end{tabular}
}
\end{table*}

\section{Prompt Templates}
\label{app:prompts}

This section documents the specific prompt templates utilized by \model. To ensure consistent behavior across heterogeneous models, we employ structured instructions for decomposition, adaptive execution, and fallback mechanisms.

\subsection{Node Decomposition (Planner)}
The \textsf{Decompose} action is executed by the Large model (30B) to generate a structured reasoning graph. It utilizes a JSON-enforced output format to ensure the router can parse subtasks and evidence.

\begin{tcolorbox}[colback=gray!5,colframe=gray!50,title=Decomposition Prompt (System \& User)]
\small
\textbf{System Prompt:} You are an expert at analyzing complex problems. Your task is to: (1) Identify key components; (2) Break it down into sub-tasks; (3) Extract evidence for each sub-task; (4) Produce concise summaries for downstream subtasks. Output a JSON object with \texttt{root\_summary}, \texttt{branch\_summary}, and \texttt{subtasks} (1--5 items). \\
\textbf{User Input:} Analyze this complex problem and break it down into manageable sub-tasks. \\
\textbf{Problem Content:} \{content\} \\
\textbf{Root Flag:} \{IS\_ROOT\}
\end{tcolorbox}

\subsection{Adaptive Execution (IO / CoT / Subtask)}
Nodes are assigned to different models based on their predicted tier: Small (4B) for \textsf{IO}, Medium (8B) for \textsf{CoT}, and Large (30B) for complex \textsf{Subtask} reasoning.

\begin{tcolorbox}[colback=gray!5,colframe=gray!50,title=Atomic Execution Templates]
\small
\textbf{Direct (Small):} Output ONLY the final answer with NO explanation, reasoning, or thinking process. Answer: \\
\textbf{CoT (Medium):} Think through this problem step by step. After your reasoning, provide your final answer in $<$answer$>$ tags. \\
\textbf{Subtask (Large):} You are solving a sub-task as part of a larger problem. \\
\textit{Context:} === ORIGINAL QUESTION === \{root\_question\} === EVIDENCE === \{planner\_evidence\} === PARENT RESULTS === \{parent\_context\} \\
\textit{Task:} \{content\}
\end{tcolorbox}

\subsection{Plan-guided Fallback (SolveWithPlan)}
When the global scheduler restricts further expansion due to budget limits, the fallback mechanism leverages the already-generated decomposition hints to provide a direct answer.

\begin{tcolorbox}[colback=gray!5,colframe=gray!50,title=Plan-guided Fallback Prompt]
\small
You are given an original question and decomposition hints. The subtasks cannot be executed due to budget limits. Use the hints to answer the original question directly. \\
\textbf{Original question:} \{question\} \\
\textbf{Summary context:} \{summary\} \\
\textbf{Decomposition hints:} \{hint\_lines\} \\
Return the final answer directly.
\end{tcolorbox}

\subsection{Final Synthesis and Evaluation Judge}
The final answer is aggregated from all leaf results, and the correctness is verified using a semantic equivalence judge implemented via Qwen3-30B.

\begin{tcolorbox}[colback=gray!5,colframe=gray!50,title=Synthesis and Judge Prompts]
\small
\textbf{Synthesis:} Synthesize the results from the following sub-tasks to answer the original question. Your LAST LINE must be "Answer: X" where X is the final answer or option letter. \\
\textbf{Semantic Judge:} Evaluate if the following two answers are semantically equivalent. Return true if they convey the same meaning, even if worded differently. Return a JSON object with \texttt{is\_equivalent} (bool) and \texttt{reasoning} (string).
\end{tcolorbox}


\end{document}